\documentclass[review]{elsarticle}

\usepackage{amsmath,color,graphicx,amssymb,float}
\usepackage{subfigure}
 \usepackage{booktabs}
 \usepackage{multirow}
 \usepackage{enumerate}
  \usepackage{tabularx}
  \usepackage{algorithm}
  \usepackage{algorithmic}
\usepackage{geometry}
\journal{Journal of \LaTeX\ Templates}
\def\x{{\mathbf x}}

\begin{document}
\begin{frontmatter}
\title{A Comparative Study of Feature Expansion Unit for 3D Point Cloud Upsampling}
\author{$\text{Qiang Li}^*$}
\cortext[mycorrespondingauthor]{Corresponding author}
\ead{liqiang19@mails.tsinghua.edu.cn}
\author{Tao Dai}
\author{Shu-Tao Xia}
\address{Tsinghua Shenzhen International Graduate School, Shenzhen, Guangdong,  China }
\begin{abstract}
Recently, deep learning methods have shown great success in 3D point cloud upsampling. Among these methods, many feature expansion units were proposed to complete point expansion at the end. In this paper, we compare various feature expansion units by both theoretical analysis and quantitative experiments. We show that most of the existing feature expansion units process each point feature independently, while ignoring the feature interaction among different points. Further, inspired by upsampling module of image super-resolution and recent success of dynamic graph CNN on point clouds, we propose a novel feature expansion units named ProEdgeShuffle. Experiments show that our proposed method can achieve considerable improvement over previous feature expansion units.
\end{abstract}
\end{frontmatter}
\section{Introduction}
\label{sec:intro}
With the continuous improvement of laser sensor, the data processed by computer has gradually expanded from images to more complex 3D forms. These 3D data are usually expressed in the format of mesh, voxel, point cloud and so on. 3D data processing is involved in many emerging scenes, such as automatic drive. Laser radar becomes necessary hardware facility to perceive the external environment and it will get surrounding three-dimensional information according to certain frequency rate constantly, while transfering the information in the form of point clouds. However, raw point clouds are often sparse, non-uniform and noisy. Thus, 3D point cloud upsampling is an important topic in 3D data processing field.

For 3D model storage, mesh data is more classic, which includes point coordinate information, edge and surface information formed by connecting multiple points, object grouping information. Since there is no regularity for edge and surface information, it is not convenient for the neural network to process. Therefore, point clouds are commonly used by deep learning methods thanks to simplicity and regular structure. Point clouds can be represented as a set of points in a three-dimensional space:
\begin{equation}
   S_N = \lbrace \mathbf{x}_1,\mathbf{x}2,\dots,\mathbf{x}_N\rbrace,
  \label{eq01}
\end{equation}
where $N$ denotes the number of points in the point cloud, and $\mathbf{x}_i$ denotes the point coordinates, which is generally a vector of length three. In the processing of point cloud data, it is generally required to have permutation invariance. Specifically, permutation invariance means that for arbitrary order of input point clouds, we should have:
\begin{equation}
  f(\mathbf{x}_1,\mathbf{x}_2,\dots,\mathbf{x}_N) \equiv f(\mathbf{x}_{\pi_1},\mathbf{x}_{\pi_2},\dots,\mathbf{x}_{\pi_N}),
  \label{eq02}
\end{equation}
where $f$ is the processing function for point cloud data, which requires that the same output can be maintained in the case of arbitrarily changing the input order of the point cloud. For example, max function, summation function and average function all meet such requirement.

In recent years, a variety of deep learning-based methods \cite{yu2018pu,yu2018ec,yifan2019patch,li2019pu,qian2020pugeo,qian2021pu} have been proposed, whose upsampling performance has significantly surpassed previous optimization-based methods \cite{alexa2003computing,lipman2007parameterization,huang2013edge,wu2015deep}. These methods propose various feature expansion units accordingly, such as schemes based on replication, latent code, Multilayer Perceptron (MLP) and graph convolution. However, most proposed feature expansion units fail to use the neighborhood information of the points, while processing each individual point in isolation, so that the generated upsampled point cloud is equivalent to independently generating several supplementary points around each point. Since there is no interaction with the features of the points in the neighborhood, it is impossible to generate supplementary points at the appropriate position based on the feature information of the surrounding points, so it can not meet the requirements for the uniformity of the generated point cloud. In addition, some methods based on the idea of graph convolution are limited by the index shape and can not perform local feature extraction on high-power point clouds.

In this paper, we firstly compare the performance of proposed feature expansion units of several classic models. To get more results, we design several feature expansion units of simple form based on MLP and shuffle structure. Then we draw on the idea of upsampling module from image super-resolution task, performing local feature fusion on high-power point clouds. Further, we proposes a step-by-step method. The feature fusion module is inspired by the success of dynamic graph CNN on point clouds. The whole feature expansion unit is called ProEdgeShuffle, and extensive experiments have shown the effectiveness of proposed feature expansion unit. In summary, the main contributions of the paper are as follows:
\begin{enumerate}
\item We fairly compare the performance of various current feature expansion units and analyze their limitations.
\item We propose a method to perform local feature fusion on high-power point features, so that we can perform point cloud upsampling and neighborhood feature interaction step by step.
\item We propose a feature expansion unit to progressively fuse local feature while expanding features, which achieves consistent performance improvements over several classic models.
\end{enumerate}

\section{Related Works}
\label{sec:related}

In the field of 3D point cloud upsampling, deep learning methods have significantly outperformed previous optimization-based methods in recent years.  PointNet \cite{qi2017pointnet} firstly applied neural networks to the task of point cloud classification and segmentation, opening a precedent for applying deep learning to point cloud data processing. Based on PonitNet, they further proposed a scheme to gradually extract local features, which greatly improved the performance of the original network. The new network is called PonitNet++ \cite{qi2017pointnet++}.

PU-Net \cite{yu2018pu} successfully applied deep networks to point cloud upsampling task firstly. They proposed to process point clouds by extracting multi-scale features, and then pass the features to a multi-branch MLP to expand the number of point clouds. The feature expansion unit of PU-Net can be considered as a multi-branch MLP. Afterwards, they proposed to promote the generation of sharper edge point clouds by introducing a loss function that is aware of side information \cite{yu2018ec}. However, this scheme requires supervised data containing side information for training. MPU\cite{yifan2019patch} proposed to gradually generate upsampled point clouds by learning features of different scales. As for feature expansion unit, inspired by the conditional image generation model \cite{mirza2014conditional}, they proposed to copy features first and then use different latent codes to map features to different spatial positions. At the end, features are merged to complete the expansion of points. PU-GAN\cite{li2019pu} introduced the idea of GAN. They designed a discriminator and added the self-attention mechanism unit\cite{zhang2019self}, which achieved state-of-the-art. PUGeo-Net \cite{qian2020pugeo} applied the idea of discrete differential geometry to deep learning networks, which can handle non-uniform distribution and peroform upsampling on noisy sparse point clouds well. Recently, DGCNN\cite{wang2019dynamic} introduced the idea of graph convolution to the processing of point cloud data and proposed a point cloud processing module EdgeConv based on graph convolution, which can obtain the local features of point clouds well. Based on the success of dynamic graph CNN, PU-GCN \cite{qian2021pu} proposed to use graph convolution for point cloud feature extraction and upsampling and achieved state-of-the-art.

Among above studies, PU-Net \cite{yu2018pu}, MPU \cite{yifan2019patch}, and PU-GCN \cite{qian2021pu} become comparison objects of many studies as for their classic structure and high performance. In this paper, we also selects these three models as base models.

 \section{Method}
 \label{sec:method}

 \subsection{Feature expansion unit}

 For 3D point cloud upsampling task, the model backbone generally includes two parts: feature extraction unit and feature expansion unit. The first part is used to extract deep features of the point cloud without changing the number of points. The feature expansion unit refers to expanding the features according to the specified upsampling ratio $r$ to achieve the purpose of multiplying the number of points. 

\begin{figure}
  \centering
  \includegraphics[width=0.4\linewidth]{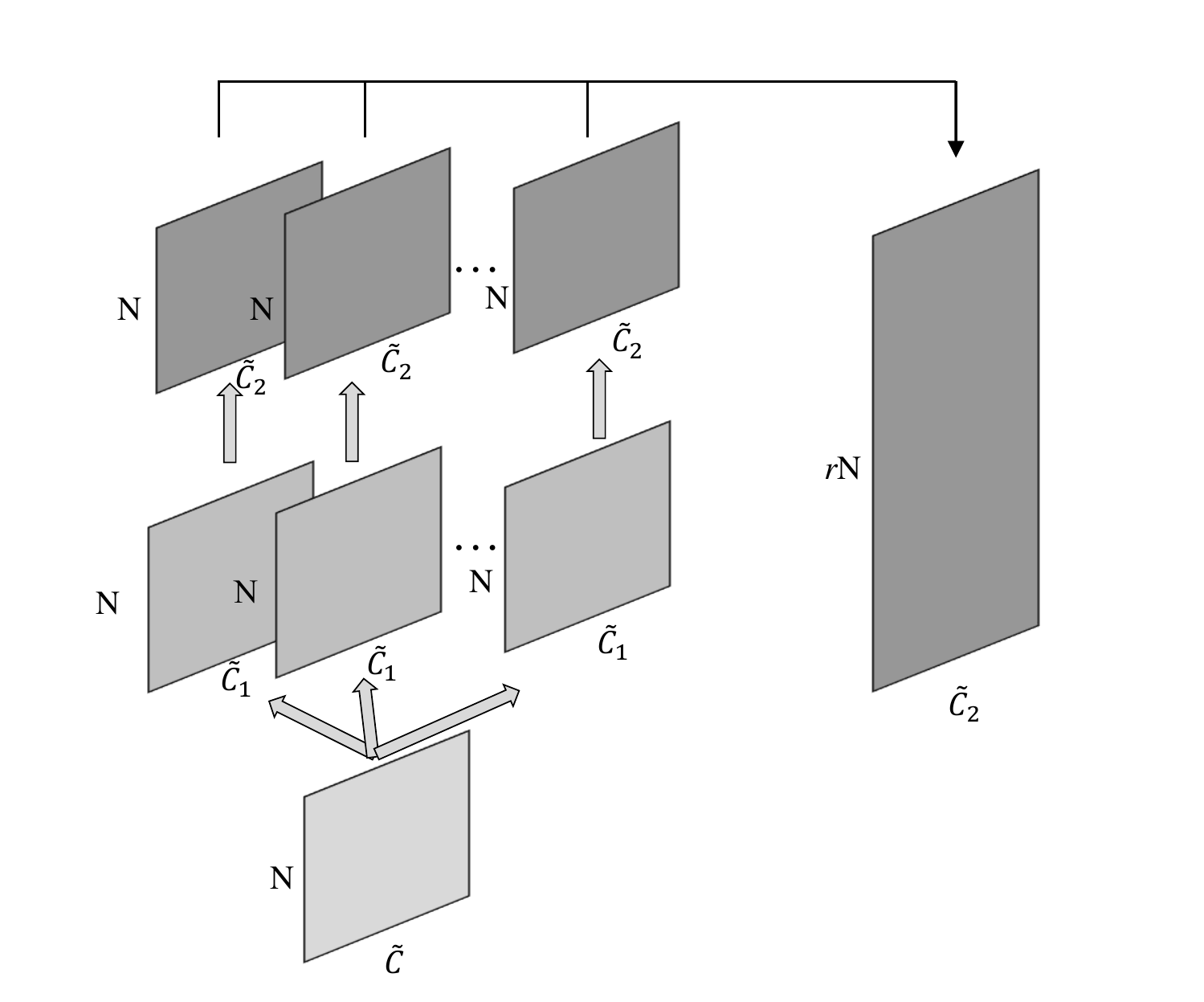}
  \caption{Feature expansion unit based on branch structure}
  \label{fig01}
\end{figure}

 Among the current point cloud upsampling models, a variety of feature expansion units have been proposed. PU-Net\cite{yu2018pu} is a pioneering work in the field of point cloud upsampling with deep learning methods. This model designs a feature expansion unit based on branch structure, as shown in Figure ~\ref{fig01}. It copies $r$ parts of features into different branches for processing, and then merges them to obtain extended features.

In general, the input of the point cloud upsampling model is $N$ three-dimensional coordinate points, and the tensor shape is $N\times3$, which becomes $N\times C$ after the feature extraction unit ($C$ is the feature dimension of a point). After the feature expansion unit, it becomes $rN\times C$, and finally $rN$ three-dimensional coordinate points are obtained through coordinate regression. As previously introduced, the point cloud upsampling module must also satisfy permutation invariance. Denote the input feature of the feature expansion unit as $\mathbf{f}$ and the output feature as $\mathbf{f^{\prime}}$, then the operation in the figure can be represented as:
\begin{equation}
  \mathbf{f^{\prime}} = H_{RS} \left( C_1^2\left(C_1^1\left(\mathbf{f}\right)\right), \dots, C_r^2\left(C_r^1\left(\mathbf{f}\right)\right) \right),
  \label{eq03}
\end{equation}
where $C_i^1$ and $C_i^2$ are both convolution operations of size $1\times1$ and all convolutions use independent parameters. $H_{RS}$ denotes reshape function. For input feature with shape $N\times\widetilde{C}$, it first copies features for $r$ times and sends them to $r$ branches. Then it uses $C_i^1$ and $C_i^2$ to transfer the dimension of features to $\widetilde{C}_1$ and $\widetilde{C}_2$. At the end, it concats these features and reshape the tensor shape to $rN\times\widetilde{C}_2$. For PU-Net, it further uses MLP to get $rN$ three-dimensional coordinate points.

\begin{figure}
  \centering
  \includegraphics[width=0.3\linewidth]{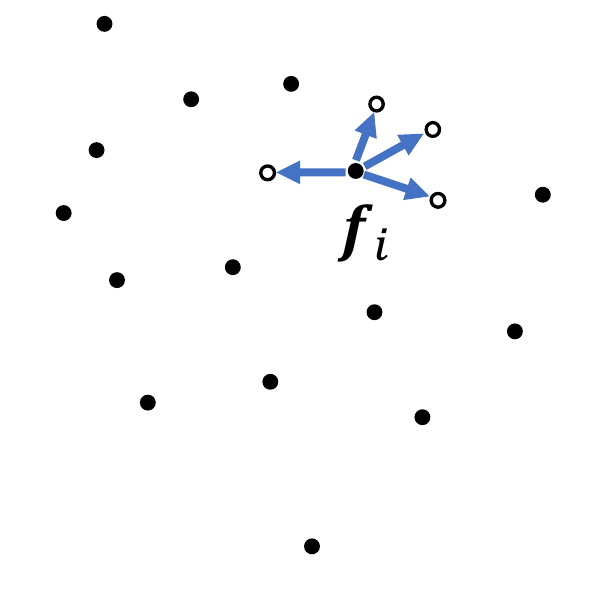}
  \caption{Illustration of branch-based feature expansion unit in feature space}
  \label{fig02}
\end{figure}

It should be noted that when performing a convolution operation with the size of $1\times1$, the range of the multiplication and addition operation of each convolution only includes the $C$ dimension vector of a single point, which means that it does not involve the feature interaction between points. In the subsequent MLP operations, only the feature dimension is reduced from $\widetilde{C}_2$ to 3 dimensions, and the feature interaction between points is also not involved. Therefore, the spatial information of the expanded point set obtained by this feature expansion unit is completely derived from the backbone network. After the feature extraction unit, there is no interaction between the features of different points. For each input point feature, it is expanded to $rC$ dimensions from its own $C$-dimensional information and then split to obtain $r$ $C$-dimensional features. The schematic diagram in feature space is shown as ~\ref{fig02}.

Let the input feature denoted as $\mathbf{f}$, which contains $N$ $C$ dimension. Each solid black dot in the figure represents the $C$ dimension feature of a single point. After feature expansion unit, a single feature $\mathbf{f}_i$ generates features of the surrounding $r$ points
relying on its own $C$-dimensional features and learned MLP parameters. The hollow points in the figure represent  generated features for $r$ points around $\mathbf{f }_i$. It can be seen that the features of any other points are not used in the generation process, so the position information of the surrounding points can not be obtained, and the points with appropriate distances can only be generated relying on priori knowledge. In this case, the generated points can not meet the requirements of uniformity.

Another design of feature expansion unit is to add a one-dimensional value after the input features are copied, so that the features can be mapped to different spaces in the subsequent processing.  This idea of adding latent code after features has been used in many models, such as image generation model conditional GAN\cite{mirza2014conditional}, point cloud autoencoder model FoldingNet\cite{yang2018foldingnet} and 3D object surface generation model AtlasNet\cite{groueix2018papier}. In the field of point cloud upsampling, MPU\cite{yifan2019patch} model designs a feature expansion unit based on this idea.

\begin{figure}
  \centering
  \includegraphics[width=0.3\linewidth]{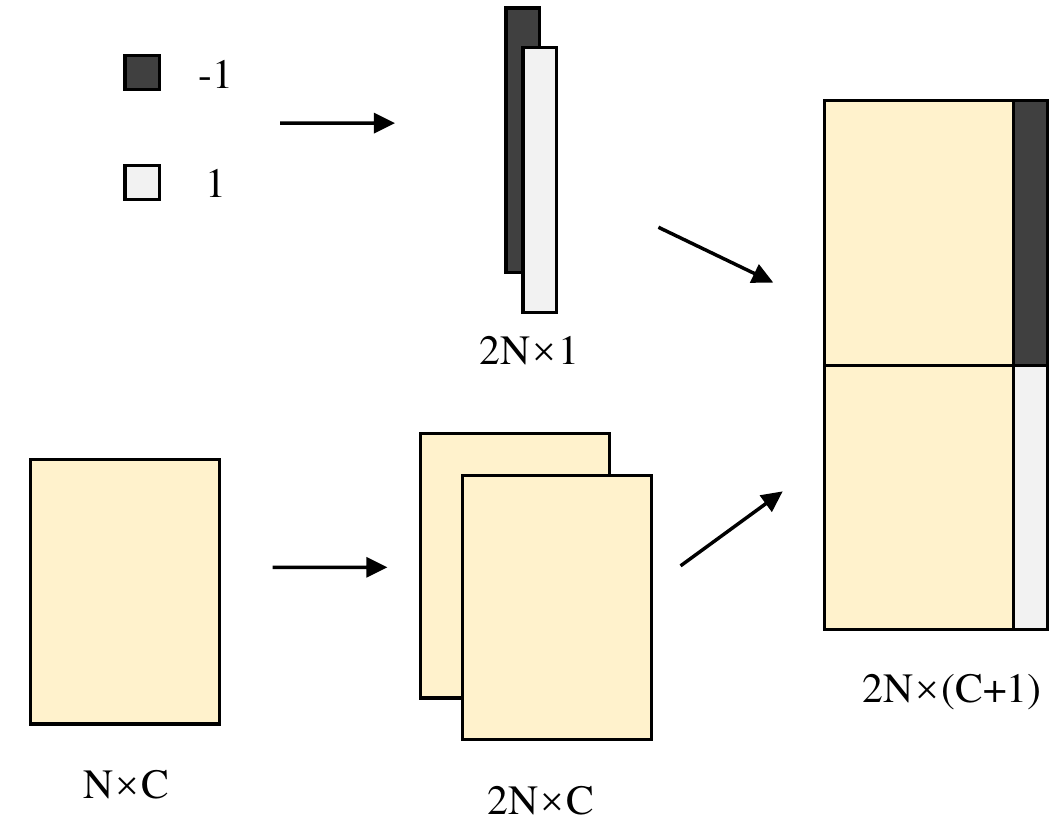}
  \caption{Feature expansion unit based on duplicate}
  \label{fig03}
\end{figure}

As shown in the figure ~\ref{fig03}, feature expansion unit based on duplicate is a module that performs $\times2$ upsampling. This module is generally used multiple times in point cloud upsampling to achieve $\times4$ and $\times16$ upsampling. It first copies the input features to get the features of shape $2N\times C$, and then splices a vector with shape $N\times1$ dimension at the end of one of them, whose values are all 1. For the other, a vector of value -1 is spliced in the same way. Then two copies are spliced together to form $2N$ $C+1$ dimension features. After the module, operations such as MLP can be used to convert the $C+1$ dimension to the specified dimension to achieve the effect of feature expansion.

\begin{figure}
  \centering
  \includegraphics[width=0.3\linewidth]{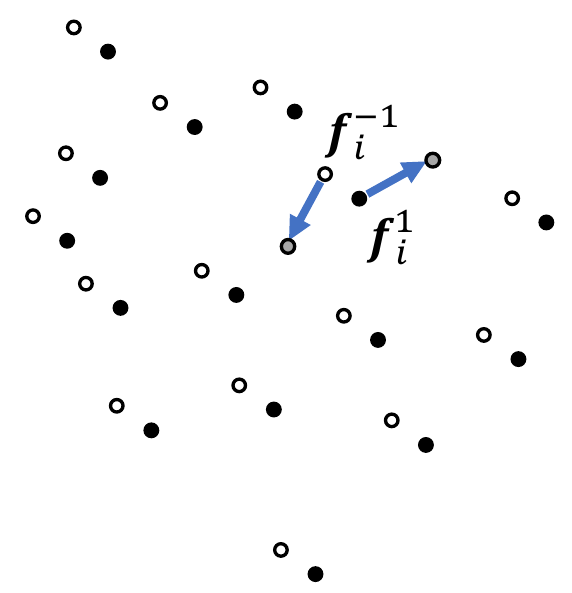}
  \caption{Illustration of duplicate-based feature expansion unit in feature space}
  \label{fig04}
\end{figure}

The original intention of this feature expansion unit is to explicitly represent the position information, so that the generated points are scattered according to different position codes, rather than simply clustered around the original points. However, by analyzing the processing flow of features, it is still found that this feature expansion unit fails to utilize the feature information of other points in the space. As shown in the figure ~\ref{fig04}, it is the schematic diagram of this feature expansion unit in the feature space. It can be seen that the only difference between two sets of features is the last dimension, so two sets of points are symmetrically distributed in the feature space. The solid black dots and the hollow dots represent the two sets of features copied respectively. The solid black dots represent the point set with a value of 1 added at the end and hollow dots represent the set of points with a value of -1 added at the end. Since the subsequent processing is generally MLP operations, it is also processed from point feature to point feature independently, which does not involve the feature interaction between points. The solid gray points shown in the figure are generated by $\mathbf{f}_i^1$ and $\mathbf{f}_i^{-1}$ independently, which means that they are processed separately. In summary, in the duplicate-based feature expansion unit, expanded features are determined by single point feature of $C$dimension, latent code and parameters of MLP.

\subsection{MLP}
\label{sec:proposed}

In the field of image processing, CNN can aggregate the local information of the picture well. As the multi-layer CNN extends, the features also transmit information in the spatial dimension. In the field of point cloud upsampling, the most commonly used module is MLP with shared parameter.

\begin{figure}
  \centering
  \includegraphics[width=0.4\linewidth]{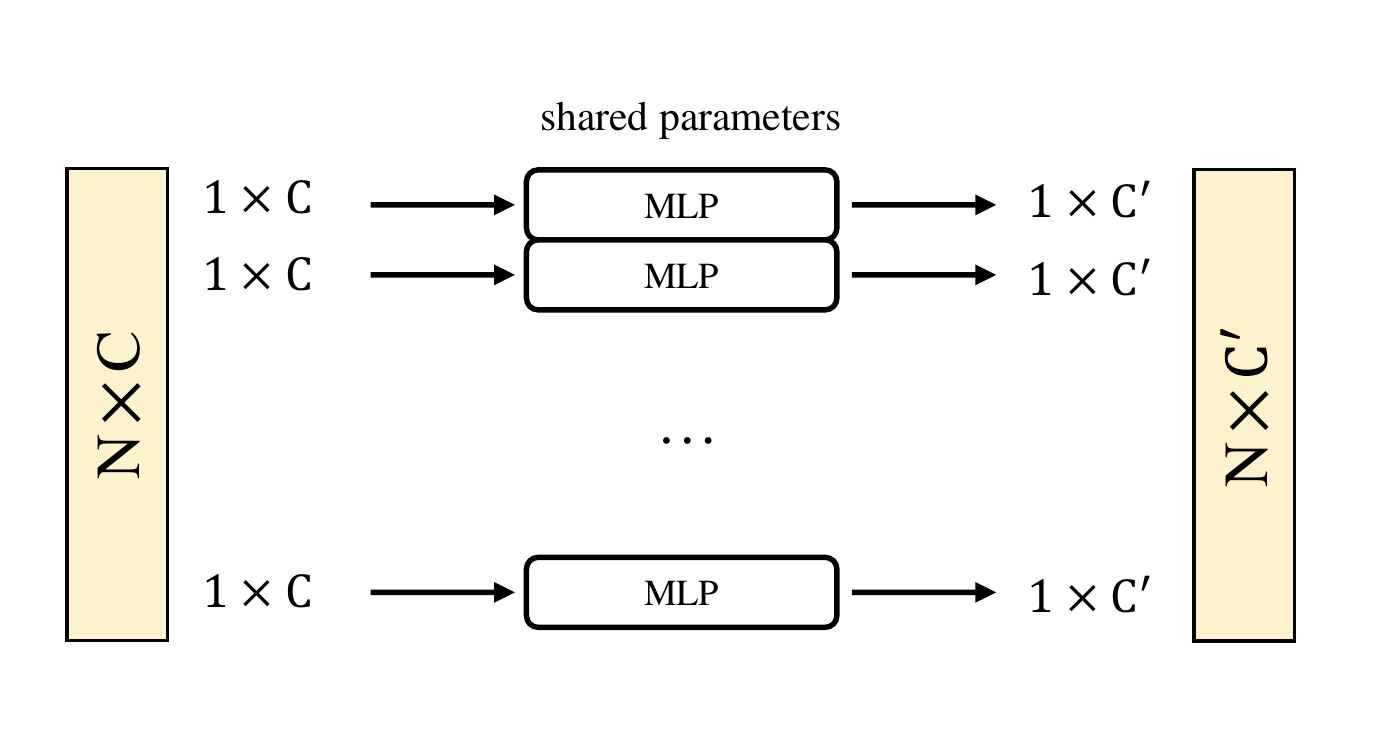}
  \caption{Illustration of MLP to extract features}
  \label{fig05}
\end{figure}

 The mechanism for MLP to process point cloud features is shown in figure ~\ref{fig05}. For a $N\times C$ input feature, the module first divides it into $N$ vectors of $C$ dimensions, and then uses MLP with shared parameters for each vector to perform feature extraction and dimension conversion. The $N$ output vectors with $C^{\prime}$ dimensions are spliced together to form the output feature with shape $N\times C^{\prime}$ finally.

It can be seen from the schematic diagram that the effect of the MLP module is only to map the features of points from one $C$ dimension feature domain to another $C^{\prime}$ dimension feature domain. The parameters of the MLP determine the way of mapping, Sharing MLP parameters is to keep same features mapping rules. In the whole process, the features between any two points do not have interaction, which is similar to an encoding and decoding process. However, the point cloud upsampling task requires that there is suitable distance between different points, and the distribution of all points should be as uniform as possible. It is difficult to achieve these requirements without interaction between different points.

\subsection{Local feature fusion method}

\begin{figure}
  \centering
  \includegraphics[width=0.4\linewidth]{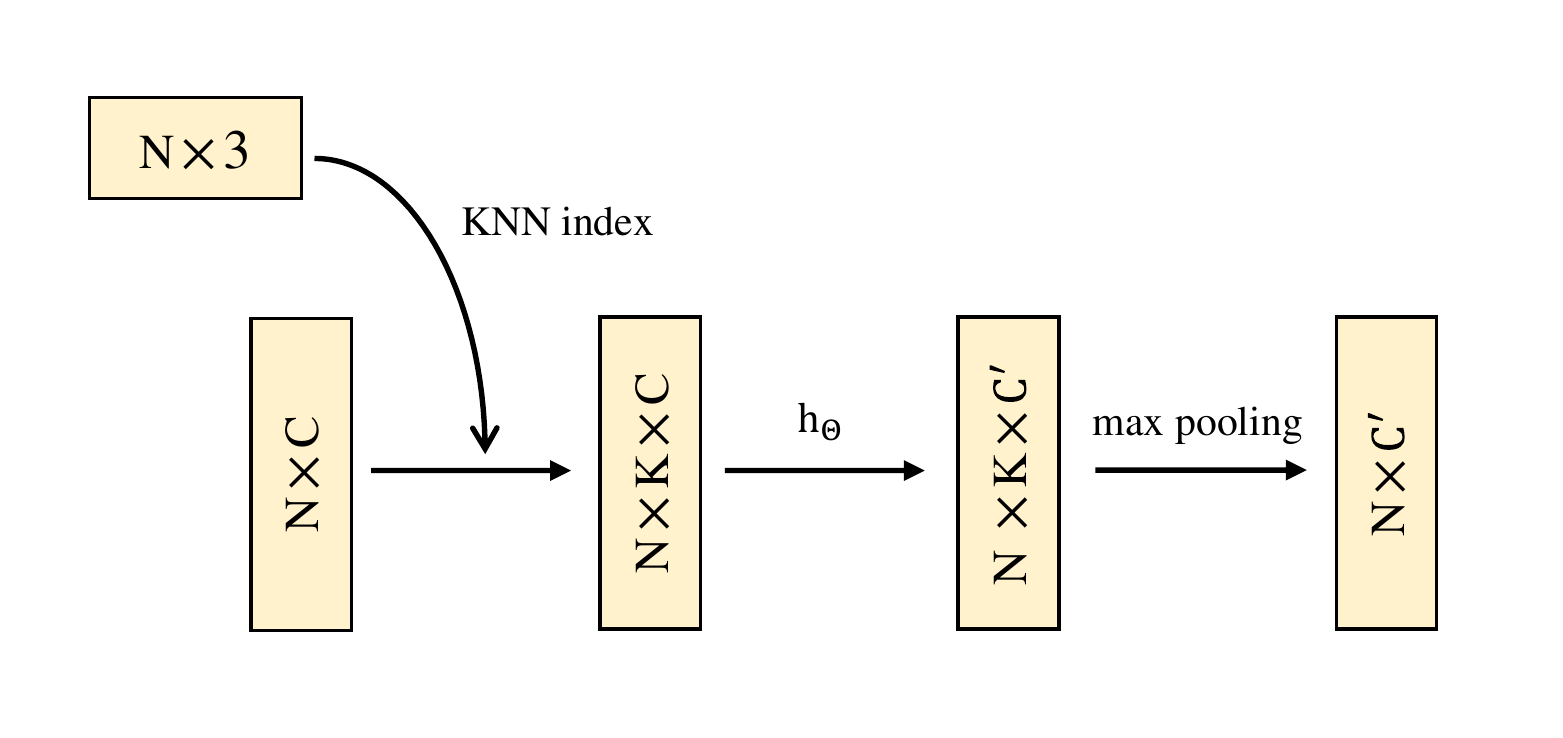}
  \caption{Illustration of EdgeConv}
  \label{fig06}
\end{figure}

For images with standard grid structures, convolutional neural networks can be used to perform local feature fusion, while for general graph structures, graph convolution can be used for feature processing and transformation. DGCNN\cite{wang2019dynamic} designed a feature extraction module for point cloud data, called EdgeConv, whose structure is shown in figure ~\ref{fig06}. It can effectively aggregates information of  neighborhood points.

For input feature with shape $N\times C$, let us denote these $N$ point features as $\mathbf{x}_i$ ($i=1,2,\dots,N$). Firstly, the K nearest neighbors of each point are calculated in Euclidean space from the initial 3D point cloud and the index information is saved. In EdgeConv, for each $\mathbf{x}_i$, we find the features of its K nearest neighbors and splice them together. Then we get features with shape $N\times K \times C$ and pass them to $h_{\ Theta}$ function to aggregate the neighborhood feature information of each point. Finally, we perform max-pooling along the dimension of $K$ added at the beginning and get output features with shape $N\times C^{\prime}$. The function of $h_{\Theta}$ is to aggregate neighborhood information and its expression can be found in DGCNN\cite{wang2019dynamic}.

Compared with the previous MLP methods to perform feature extraction and dimension conversion, EdgeConv continuously updates the features of the current point through the feature information of neighborhood points, so that the information of each point spreads in the spatial domain, which is conducive to generating more uniform point clouds. In addition, this method meets the requirement of permutation invariance. Based on this idea of graph convolution, PU-GCN\cite{wang2019dynamic} proposed a NodeShuffle module to perform feature expansion. It firstly uses EdgeConv to perform a local feature fusion operation on the input tensor. At the same time it expands the feature dimension by $r$ times. Then it realizes the expansion of the number of points by shuffle operation, which transits the value of the feature dimension to the point dimension. Experiments on PU-GCN model show that NodeShuffle structure outperforms previous branch-based and duplicate-based modules.

\subsection{ProEdgeShuffle}

\begin{figure}
  \centering
  \includegraphics[width=0.6\linewidth]{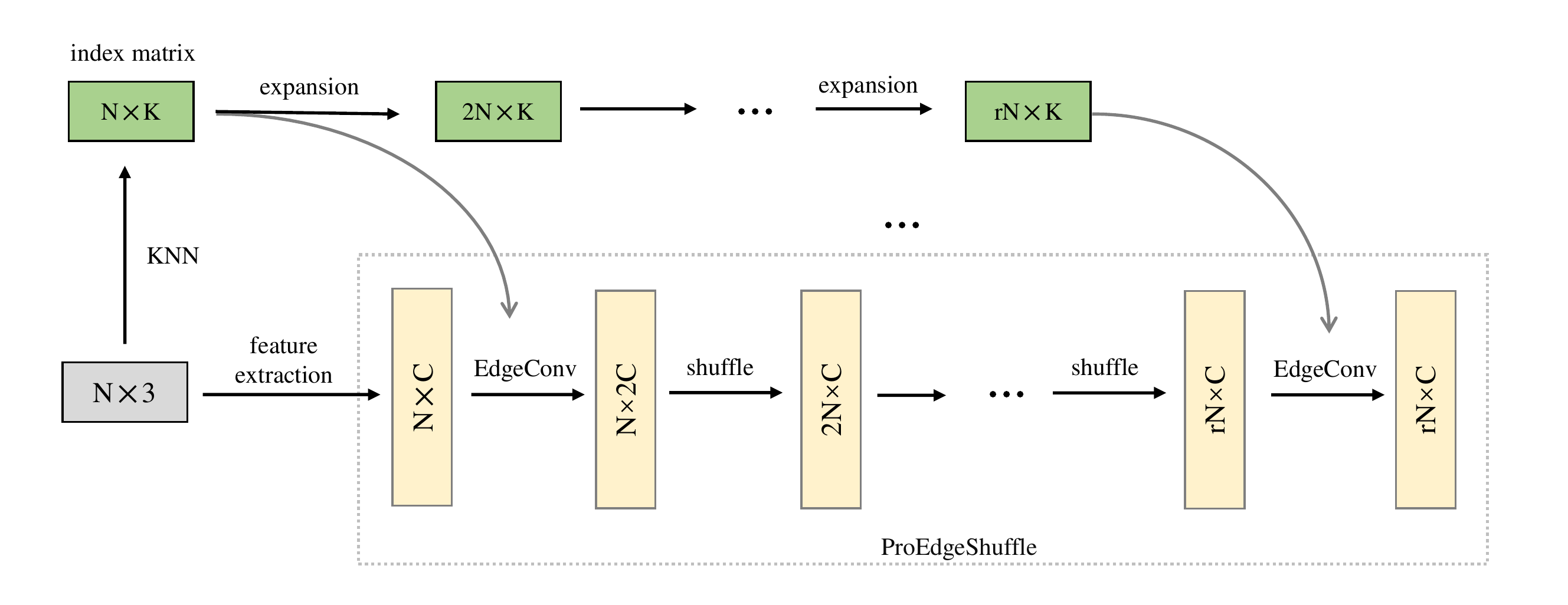}
  \caption{Illustration of ProEdgeShuffle}
  \label{fig07}
\end{figure}

Based on the idea of graph convolution, we further design a feature expansion unit called ProEdgeShuffle that gradually fuses local features. The structure is shown in the figure ~\ref{fig07}. For the point cloud input with shape $N\times3$, we firstly use a neural network model to extract its features whose shape is $N\times C$. At the same time we calculate $K$ nearest neighbor points of each point in the three-dimensional Euclidean space and save index information in index matrix. It can be considered as constructing an edge between each point and its $K$ neighbors, thus forming a graph. This graph structure keep fixed throughout the model calculation.

\begin{figure}
  \centering
  \includegraphics[width=0.4\linewidth]{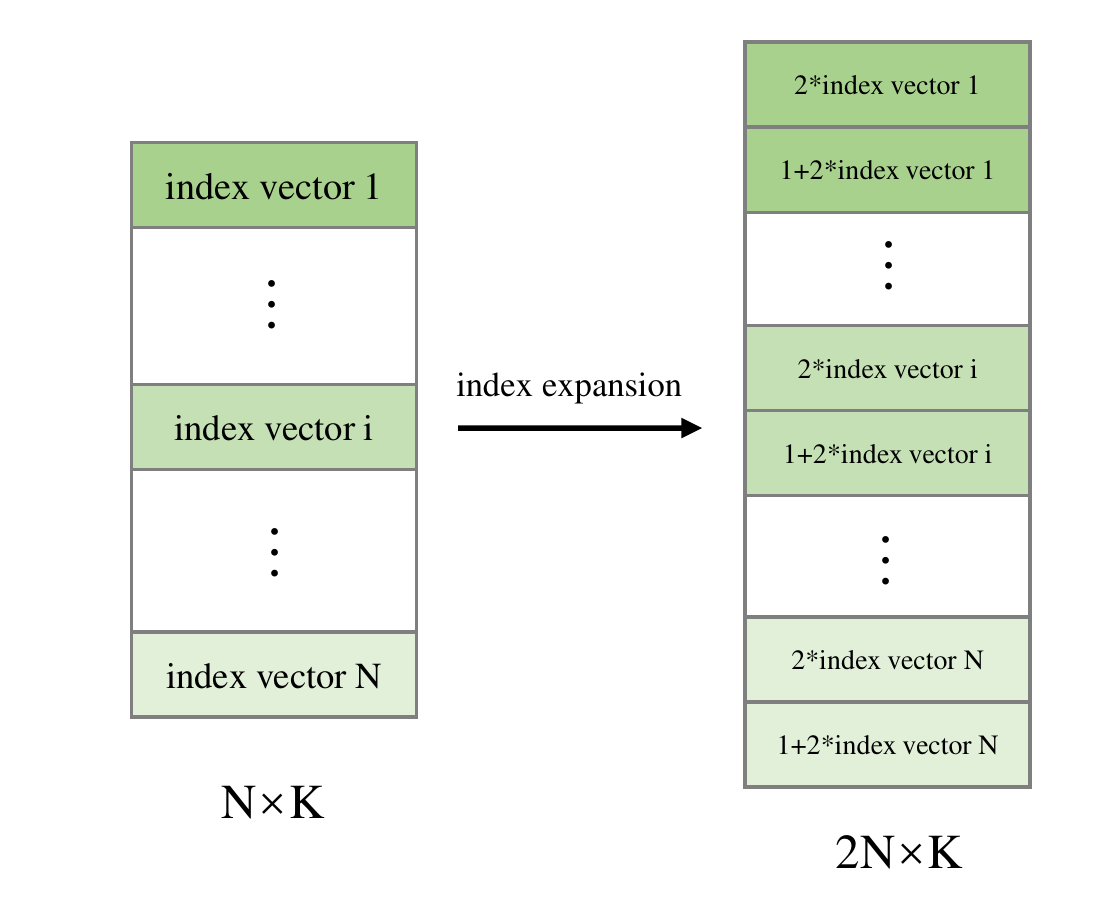}
  \caption{Illustration of index expansion}
  \label{fig08}
\end{figure}

In the processing of multiple upsampling, the branch-based method can use $r$ branches, and the duplicate-based method can increase the dimension of added hidden code or perform multiple upsampling step by step. However, for the EdgeConv, it is limited by the shape of index matrix, thus it is impossible to continue to use the EdgeConv operation when the number of points exceeds $N$. For example, after NodeShuffle performs $\times r$ feature expansion, it is impossible to perform local feature fusion again. In PU-GCN, it uses MLP to perform feature dimension conversion and coordinate regression, thus the neighborhood information between the points cannot be obtained in high-power feature space.To tackle this problem,we propose an index expansion method, as illustrated in the figure ~\ref{fig08}. After EdgeConv with $\times 2$ upsampling, the point feature at position $i$ is expand to high-power feature at position $2i-1$ and $2i$. Therefore, for the index vector at position $i$, we expand them to target index matrix at position $2i-1$ and $2i$. Accordingly, the index value should be multiplied by 2. For example, a point feature $\mathbf{f}_i$ generates a new point feature $\mathbf{f}_i^{\prime}$ around it, then they should have similar neighbor index from original index vector of $\mathbf{f}_i$.

Back to the mechanism of ProEdgeShuffle, we get high-power index matrix now. For the features extracted from feature extraction unit, whose shape is $N\times C$, we use one EdgeConv to get features with shape $N\times 2C$ and then shuffle it to get high-power feature with shape $2N\times C$. Then we use expanded index matrix to perform another EdgeConv and shuffle again to get high-power feature with shape $4N\times C$. Similarly, we can get high-power feature with shape $16N\times C$. When obtaining the feature output with shape $rN\times C$, most models usually directly use the MLP operation to convert the dimension from the $C$ dimension to 3 dimensions without additional processing. We draws on the idea of upsampling module of image super-resolution and propose to perform another local feature fusion operation. After the image super-resolution model obtains high-resolution features, it will perform a convolution operation on the high-resolution features to fuse local features, but the MLP in the point cloud can not aggregate local features. Thus, ProEdgeShuffle continues to perform an EdgeConv operation on output features with shape $rN\times C$ , which further aggregates local features on high-power feature space to make the distribution of output point cloud more uniform.

\section{ Experiments}
\label{sec:result}

\subsection*{A.  Setup}

All experiments in this paper are carried out based on the same environment setup. We conduct all experiments on Ubuntu 16.04 and NVIDIA GeForce RTX 2080 Ti and the corresponding CUDA version is 10.1. In addition, we use Python 3.6 and TensorFlow 1.13.

In terms of datasets, we use PU1K \cite{wang2019dynamic} dataset for training and testing. Recnet point cloud upsampling models use a wide variety of datasets since PU-Net first trys the neural network method. For example, PU-Net uses 60 self-collected 3D models, PU-GAN\cite{li2019pu} uses 147 self-collected 3D models, and MPU\cite{yifan2019patch} uses the existing MNIST-CP, Sketchfab and ModelNet10\cite{wu20153d} datasets. However, these datasets requires that we firstly construct the input sparse point cloud data and GT data on the 3D model using Poisson disk sampling. Poisson disk sampling output random sampling results, so the results of each test are different. Recently, PU-GCN\cite{wang2019dynamic} proposed  PU1K dataset, which collectes 1147 3D models and the scenes are very rich. At the same time, PU1K gives the standard input and GT data for testing. Thus the quantitative results of each test are the same. To facilitate comparison experiments, all experiments in this paper are performed based on PU1K dataset.

As for model training, we select three classical models, PU-Net\cite{yu2018pu}, MPU\cite{yifan2019patch} and PU-GCN\cite{wang2019dynamic}, as basic models, and we trian all three models using the same training setup, which follows the configuration of PU-GCN. Specifically, the training optimizer adopts the ADAM\cite{kingma2014adam} optimizer and the optimizer parameters are all set to default parameters except specified. The initial learning rate is 0.001, and the value of $\beta$ is 0.9. After 100 epochs of training, the training is ended and the final model is taken for testing. In the test phase, we select commonly used CD, HD and P2F for quantitative comparation. Besides, we will show the comparison of the upsampling point cloud results for qualitative comparison.

\subsection*{B. Comparison of different feature expansion units}

In this section, we compare a variety of feature expansion units. The first part is feature expansion units proposed by previous state-of-the-art models, such as the branch-based feature expansion unit in PU-Net, duplicate-based feature expansion unit in MPU and NodeShuffle in PU-GCN. The second part is some simple structure we design based on MLP and shuffle structure, which we use as a reference for comparison.Specifically, we design three types of MLP-based feature expansion units:
\begin{itemize}
  \item Single MLP-based feature expansion unit. For input features with shape $N\times C$, we firstly use a single-layer MLP (different points share MLP parameters, the same below) to convert its feature dimension from $C$ to $rC$ and then shuffle features to get shape $rN\times C$.
  \item Multilayer MLP-based feature expansion unit. For input features with shape $N\times C$, we firstly use five MLP layers to extract features. Note that output features still have the shape $N\times C$. Then we  use a single-layer MLP to convert its feature dimension to $rC$ and then shuffle features to get shape $rN\times C$.
  \item Progressive MLP-based feature expansion unit.For input features with shape $N\times C$, firstly we use a single-layer MLP for feature extraction. Then we perform single-layer MLP to double feature dimension and shuffle features cyclically. When getting features with shape  $rN\times C$, we stop the loop.
\end{itemize}

We compare the performance of these six feature expansion units on three basic models. The results are shown in the table ~\ref{table1}. For CD, HD and P2F, the smaller, the better the performance is. And the best performance is in bold. Among the six feature expansion units, it can be seen that the performance of the branch-based, duplicate-based and single MLP-based feature expansion units is not good. Possible reason is that these types of feature expansion units have relatively simple structures and do not take into account information exchange between point features. Besides, there is not enough MLP parameters to support it to learn richer upsampling mode, leading to poor performance. In contrast, multilayer MLP-based, 
progressive MLP-based and NodeShuffle feature expansion units have better performance on each model. In particular, NodeShuffle module is the best or very close to the best based on all three models. Although multilayer MLP-based and structure and progressive MLP-based structure have rich parameters, they are still essentially mapping operations on the feature space and do not involve feature interactions between points. On the contrary, NodeShuffle module builds a neighbor graph structure from point cloud information and performs interacts with the point features of their neighbors. The performance of NodeShuffle shows that local feature fusion is beneficial. As for different models, neither the PU-Net nor the MPU model network has the structure of constructing the nearest neighbor graph from the original point cloud. Using the NodeShuffle  module can greatly improve the performance. In contrast, the PU-GCN model contains a large number of graphs Convolution operation, the gain of local feature fusion from feature expansion unit is not obvious.

\begin{table}
  \centering
  \caption{Quantitative comparation of different feature expansion units}
  \begin{tabular}{ccccc}
    \toprule
      Model&Feature expansion unit&CD\ ($10^{-3}$)&HD\ ($10^{-3}$)&P2F\ ($10^{-3}$)\\
    \midrule
      \multirow{6}{*}{PU-Net}&Branch-based&1.120&14.905&4.854 \\
      &Duplicate-based&1.111&14.341&5.115 \\
      &NodeShuffle&1.004&\textbf{12.822}&\textbf{4.752} \\
      &Single MLP-based&1.073&14.979&5.111 \\
      &Multilayer MLP-based&1.037&14.043&5.212 \\
      &Progressive MLP-based&\textbf{0.981}&14.025&5.112 \\ \hline
      \multirow{5}{*}{MPU}&duplicate-based&0.960&12.794&3.430 \\
      &NodeShuffle&\textbf{0.803}&\textbf{11.251}&\textbf{3.310} \\
      &Single MLP-based&0.929&12.730&3.489 \\
      &Multilayer MLP-based&0.884&12.923&3.521 \\
      &Progressive MLP-based&0.927&13.582&3.373 \\ \hline
      \multirow{5}{*}{PU-GCN}&duplicate-based&0.692&10.148&2.566 \\
      &NodeShuffle&0.657&10.214&2.663 \\
      &Single MLP-based&0.741&11.421&2.661 \\
      &Multilayer MLP-based&\textbf{0.655}&\textbf{9.408}&2.578 \\
      &Progressive MLP-based&0.659&9.697&\textbf{2.519} \\
    \bottomrule
  \end{tabular}
  \label{table1}
\end{table}

\begin{figure}
  \centering
    \includegraphics[width=0.8\linewidth]{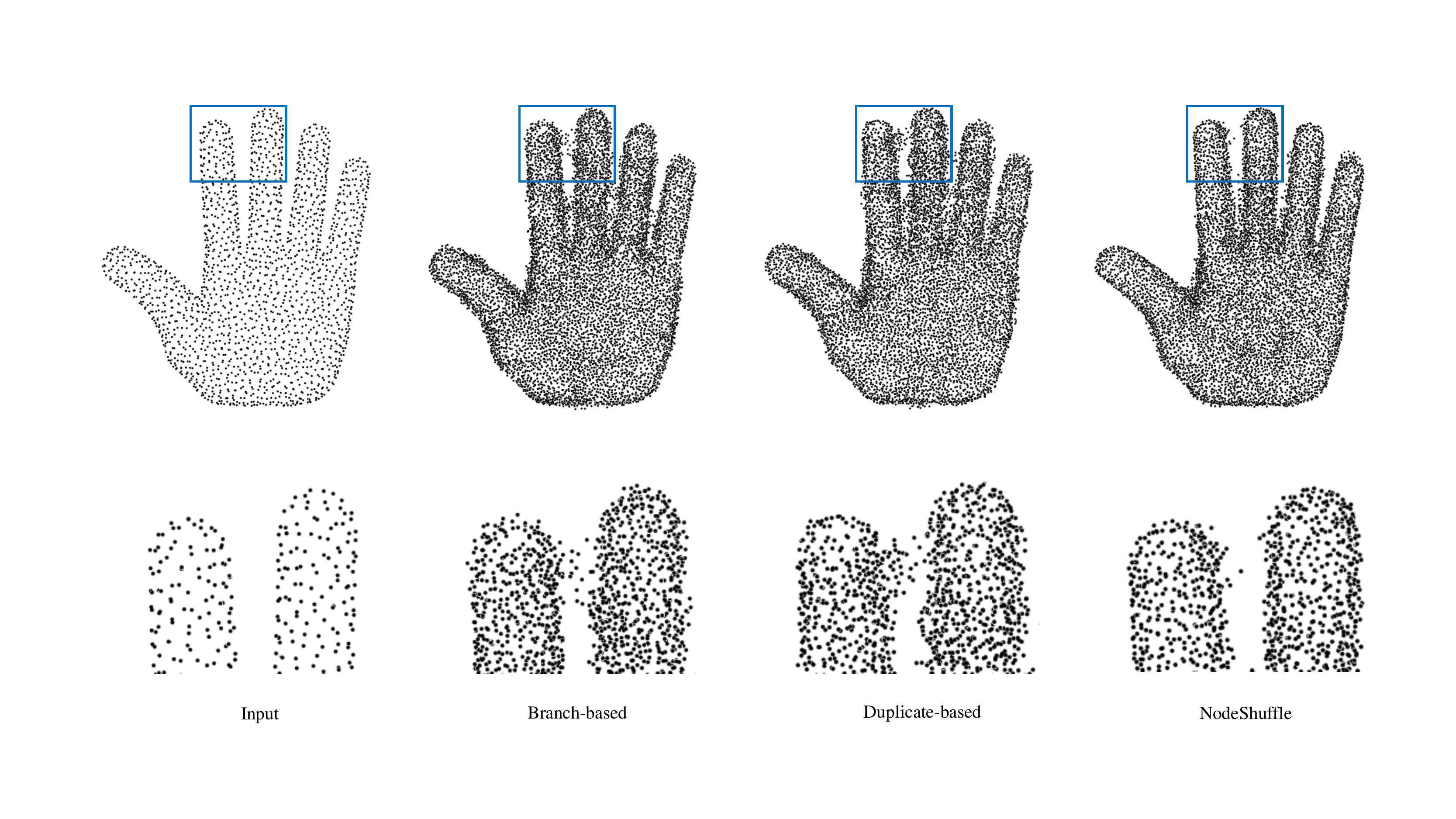}
    \includegraphics[width=0.8\linewidth]{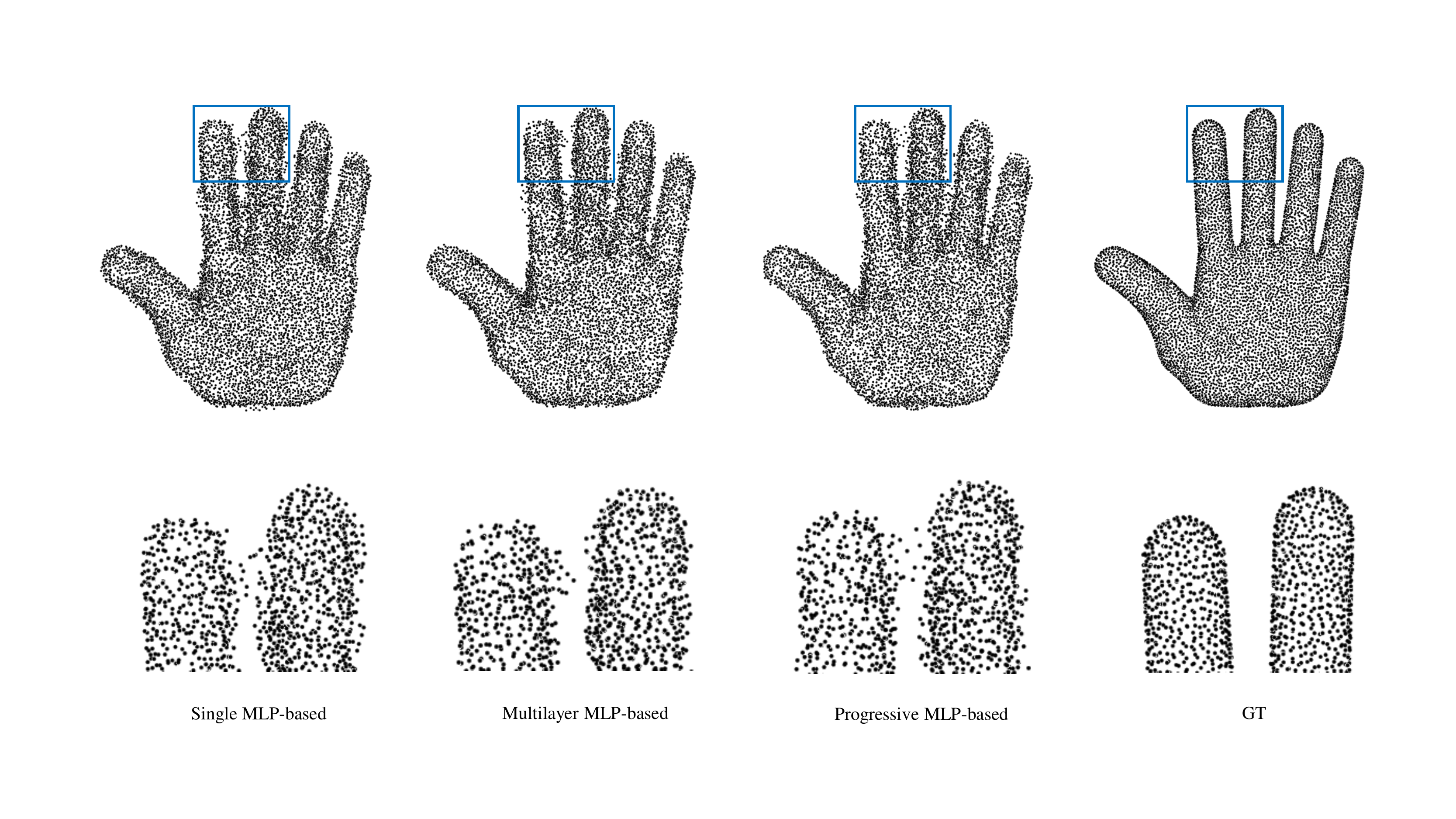}
  \caption{Visualization of different feature expansion units}
  \label{fig09}
\end{figure}

We further show the visualization results of applying feature expansion units, as shown in the figure ~\ref{fig09}. The basic model selected in the figure is PU-Net and the test object is a palm 3D model. We choose to enlarge the upper half of the index finger and middle finger. On the one hand, we can observe the uniformity of the point cloud distribution generated by the model, on the other hand, we can observe how the model generates outliers at corner seam locations. The eight results in the figure are sparse point cloud input, branch-based, duplicate-based upsampling, NodeShuffle, single MLP-based, multilayer MLP-based, progressive MLP-based feature expansion unit and GT. In terms of the uniformity of point cloud distribution, since the basic model is the early PU-Net, the generated point cloud distribution is relatively scattered, which is prone to local aggregation and small-scale blanks. However, the NodeShuffle module, multi-layer MLP and progressive MLP perform relatively well. In terms of outliers, early branch-based and duplicate-based modules generate many outliers at blank seams. Some even form sparse spaces in small areas, which hindered the recognition of 3D model contours. The three methods based on MLP are relatively better. Some outliers are still generated but they will not be misidentified. NodeShuffle module is obviously better in the processing of outliers. Only a small number of outliers are generated and the overall outline is relatively clear. The reason is that among these modules, only NodeShuffle module takes into account of the feature interaction between points and effectively receives information from neighborhood points. Thus it is difficult for NodeShuffle to generate large-scale outliers.

\subsection*{C. Results of ProEdgeShuffle}

We test the effectiveness of the proposed ProEdgeShuffle module on three base models. For the baseline experiments, three models all use the original network architecture. In contrast, for our models, we change original feature expansion unit to ProEdgeShuffle module for testing. The quantitative results on the $\times4$ point cloud upsampling task are shown in the table~\ref{table2}. The best performance of CD, HD and P2F are in bold.

\begin{table}
  \centering
  \caption{Quantitative results of ProEdgeShuffle}
  \begin{tabular}{ccccc}
    \toprule
      Model&Feature expansion unit&CD\ ($10^{-3}$)&HD\ ($10^{-3}$)&P2F\ ($10^{-3}$)\\
    \midrule
      \multirow{2}{*}{PU-Net}&Original&1.120&14.905&4.854 \\
      &ProEdgeShuffle&\textbf{0.938}&\textbf{14.512}&\textbf{3.831}\\ \hline
      \multirow{2}{*}{MPU}&Original&0.960&12.794&3.430 \\
      &ProEdgeShuffle&\textbf{0.603}&\textbf{9.102}&\textbf{2.620}\\ \hline
      \multirow{2}{*}{PU-GCN}&Original&0.657&10.214&2.663 \\
      &ProEdgeShuffle&\textbf{0.597}&\textbf{8.138}&\textbf{2.295}\\
    \bottomrule
  \end{tabular}
  \label{table2}
\end{table}

The original feature expansion units of the three models in the table are branch-based , duplicate-based and NodeShuffle module respectively. It can be clearly seen from the table that after replacing the original ufeature expansion unit with the ProEdgeShuffle, the performance of three model have been significantly improved. The improvement on the MPU is particularly obvious. CD is reduced by 37\% and HD is reduced by 29\%. There is no local feature extraction module in the PU-Net structure, and only spatial information is fused by downsampling and upsampling. In this case, the performance can be greatly improved by adding a local feature fusion module at the end of the feature expansion module. MPU structure includes K-nearest neighbor operations on some feature spaces, but there is no K-nearest neighbor information obtained from the original 3D point cloud Euclidean space. ProEdgeShuffle module makes up for this part of the information and greatly reduces the CD, HD and P2F. PU-GCN structure already contains operations similar to graph convolution, which effectively integrates local information, but ProEdgeShuffle provides local feature fusion operations on high-power point cloud feature space, which has a certain gain in model performance, so we can see a small performance gain on PU-GCN with ProEdgeShuffle.

\begin{figure}
  \centering
  \includegraphics[width=0.8\linewidth]{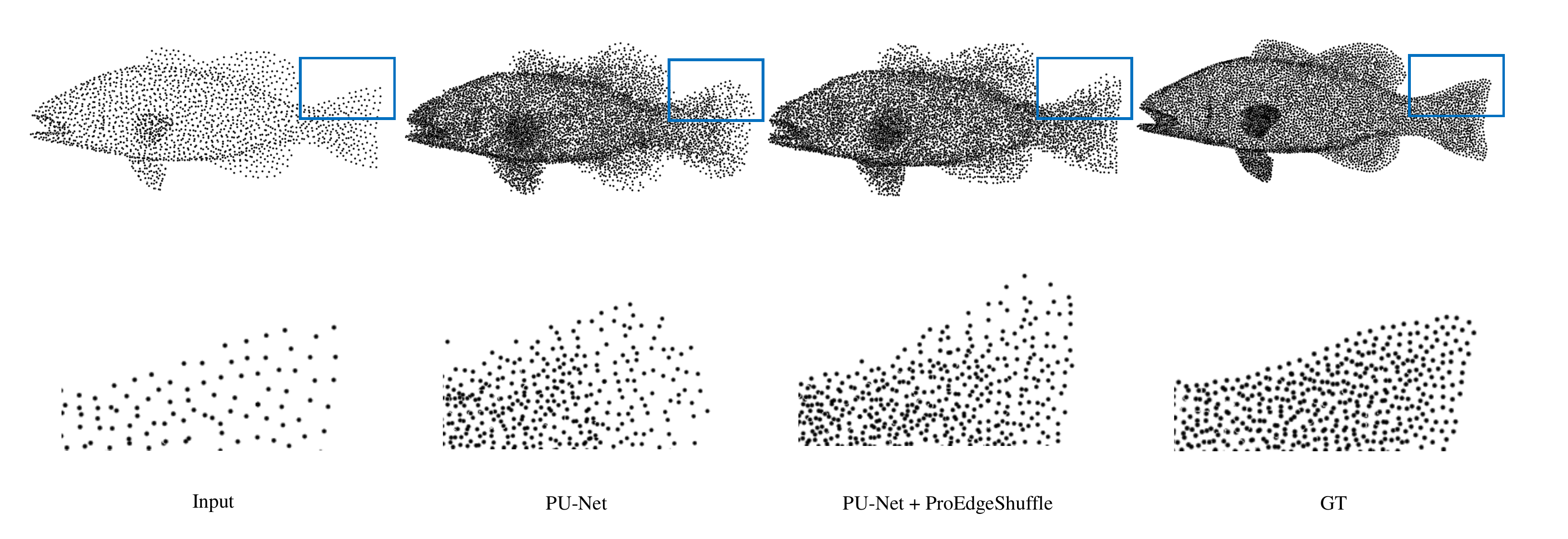}
  \caption{Visualization of ProEdgeShuffle on PU-Net}
  \label{fig10}
\end{figure}

Then we show the visualization of the ProEdgeShuffle module on three models. PU-Net is an early model of point cloud upsampling task. The visualization of the ProEdgeShuffle module applied to PU-Net is shown in the figure ~\ref{fig10}. We choose 3D fish model for visualization. It can be seen that the input sparse point cloud and the GT point cloud distribution are relatively uniform, but the input point cloud lose detailed information on the shape due to the few number of points in local position. We enlarge the space at the fishtail. The original PU-Net model loses the shape of the fishtail after upsampling, and the edge of the entire point cloud distribution is approximately arc-shaped, while the model with ProEdgeShuffle restores the approximate shape of the fishtail. Compared to the results of the original model, there are already slightly more accurate sharp-angled contours. Due to the fact that the early PU-Net model is not mature enough in feature extraction, there is still a lot of gap between the upsampling result and GT.

\begin{figure}
  \centering
  \includegraphics[width=0.8\linewidth]{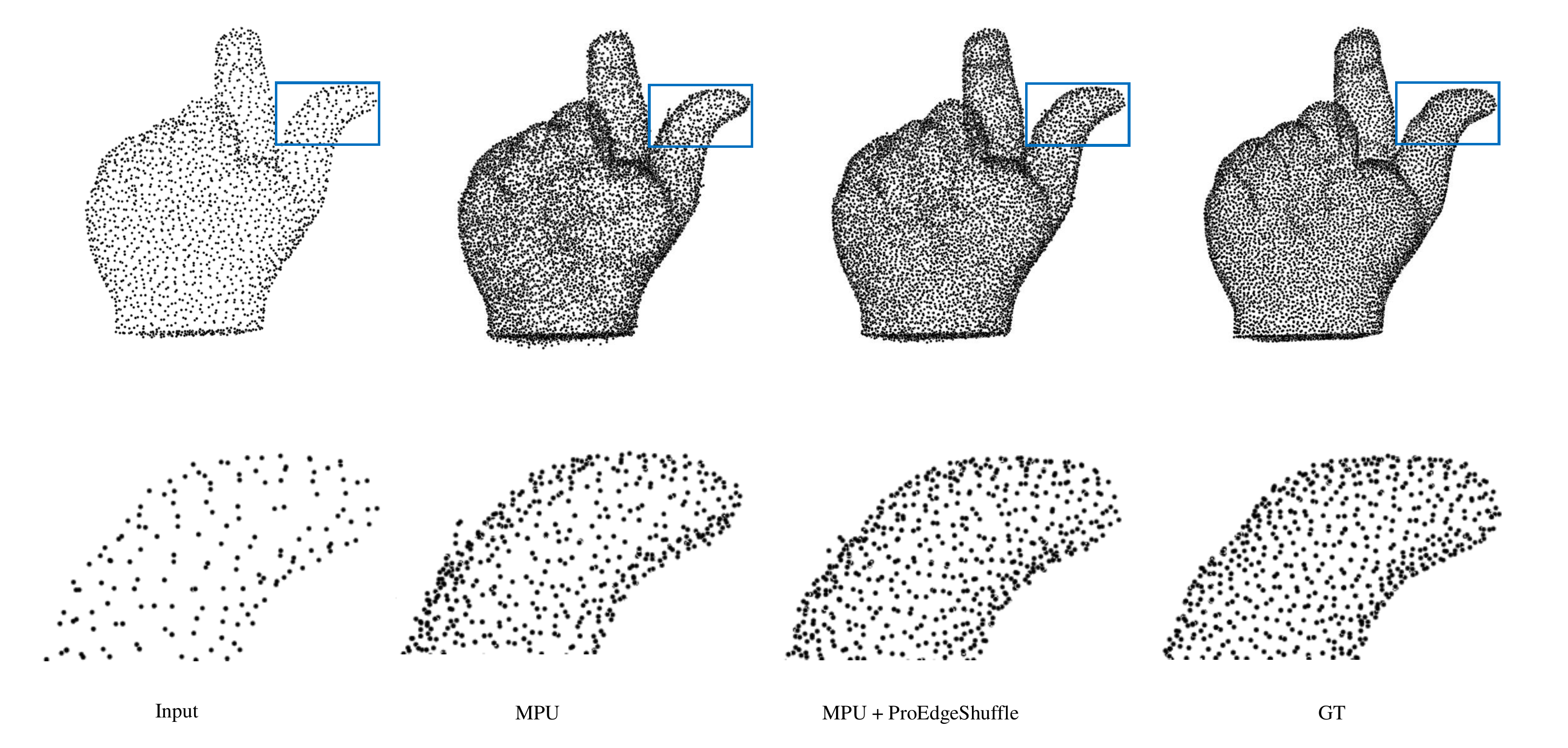}
  \caption{Visualization of ProEdgeShuffle on MPU}
  \label{fig11}
\end{figure}

\begin{figure}
  \centering
  \includegraphics[width=0.8\linewidth]{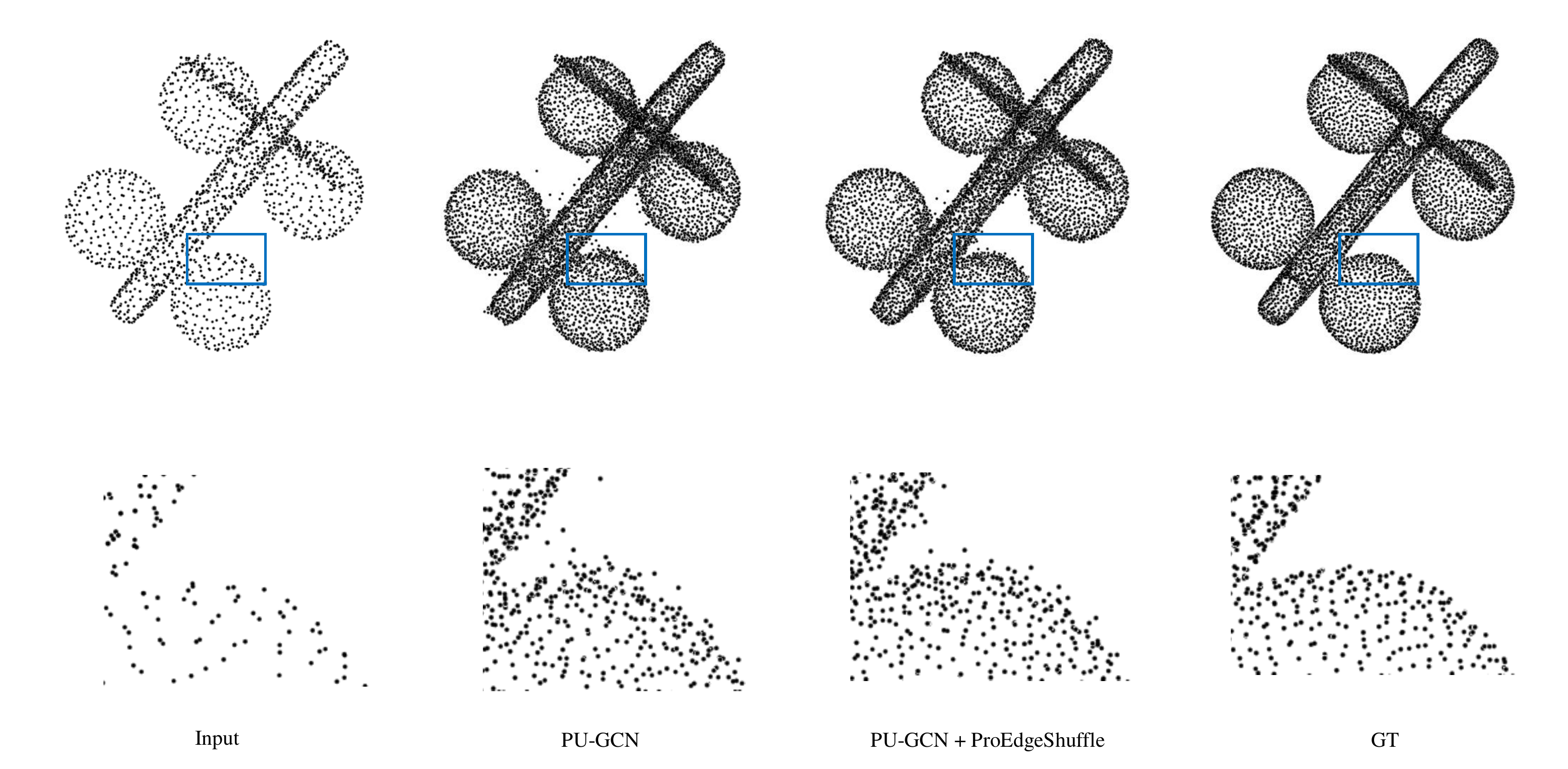}
  \caption{Visualization of ProEdgeShuffle on PU-GCN}
  \label{fig12}
\end{figure}

The visualization of ProEdgeShuffle on MPU model is shown in the figure ~\ref{fig11}, and the visualization of ProEdgeShuffle on PU-GCN model is shown in the figure ~\ref{fig12}. For MPU, a gesture model is selected. Compared with early PU-Net model, it can be seen that the point cloud recovered by MPU is close to GT in overall outline, but in terms of local details and uniformity of point cloud distribution, there are still some gaps. We zoom in the spatial position of the thumb. It can be seen that the original MPU model will generate some outliers, which are added by mistake after the model is processed. In contrast, for ProEdgeShuffle, there are frequent feature interactions with its neighbors in the process of generating points, so some points that deviate from the original model are not generated, which also confirms the effectiveness of ProEdgeShuffle.

PU-GCN is a very advanced model at present, and it is very close to GT in the recovery of the overall model outline, but some outliers are still generated in some corner positions of the 3D model. As shown in the enlarged space in the figure, this corner should be a blank part in GT. However, because there are surface points distributed on both sides, it is very easy to fill it with new points during the upsampling process. PU-GCN will be misled and generate some outliers in the corners, but the model applying ProEdgeShuffle module will interact with its original neighbor features at each step of the step-by-step upsampling and continuously update the feature information of the neighbors. So generated points will not deviate too much. It can also be clearly seen in the visualization figure that PU-GCN with ProEdgeShuffle generates fewer outliers and the distance of the outliers deviating from the main 3D model is also relatively shorter.

\subsection*{D. Ablation study}

We firstly compare two types generate high-power index.Performing EdgeConv operations on high-power features requires high-power index matrix. The first type is to recalculate K-nearest neighbors on high-power features each time  when EdgeConv is used. Then we can use the K-nearest neighbor information in the feature space toperform EdgeConv operations. The second is to use the point cloud K-nearest neighbor expansion method we propose in this paper. That means the entire model only uses the K-nearest neighbor index in the original 3D point cloud space. The expansion operation is shown in the figure ~\ref{fig08}. The quantitative comparison is shown in the table ~\ref{table3}. It can be seen that on three base models, ProEdgeShuffle using point cloud K-nearest neighbor expansion method is better for CD and HD , and the improvement is very obvious.  As for P2F, the possible explanation is that the K-nearest neighbors on the feature space can make the generated points closer but the point cloud K-nearest neighbor expansion method makes generated points closer to the original input points. So there is a slight difference for P2F. On the whole, using the point cloud K-nearest neighbor expansion method can get better quantitative results.

\begin{table}
  \centering
  \caption{Ablation study on high-power index type}
  \begin{tabular}{ccccc}
    \toprule
      Model&High-power index type&CD\ ($10^{-3}$)&HD\ ($10^{-3}$)&P2F\ ($10^{-3}$)\\
    \midrule
      \multirow{2}{*}{PU-Net}&KNN on features&1.032&15.232&\textbf{3.485} \\ 
      &Index expansion&\textbf{0.938}&\textbf{14.512}&3.831\\ \hline
      \multirow{2}{*}{MPU}&KNN on features&0.870&11.564&\textbf{2.465} \\ 
      &Index expansion&\textbf{0.603}&\textbf{9.102}&2.620\\ \hline
      \multirow{2}{*}{PU-GCN}&KNN on features&0.714&9.981&2.938 \\ 
      &Index expansion&\textbf{0.597}&\textbf{8.138}&\textbf{2.295}\\
    \bottomrule
  \end{tabular}
  \label{table3}
\end{table}

\begin{figure}
  \centering
  \includegraphics[width=0.8\linewidth]{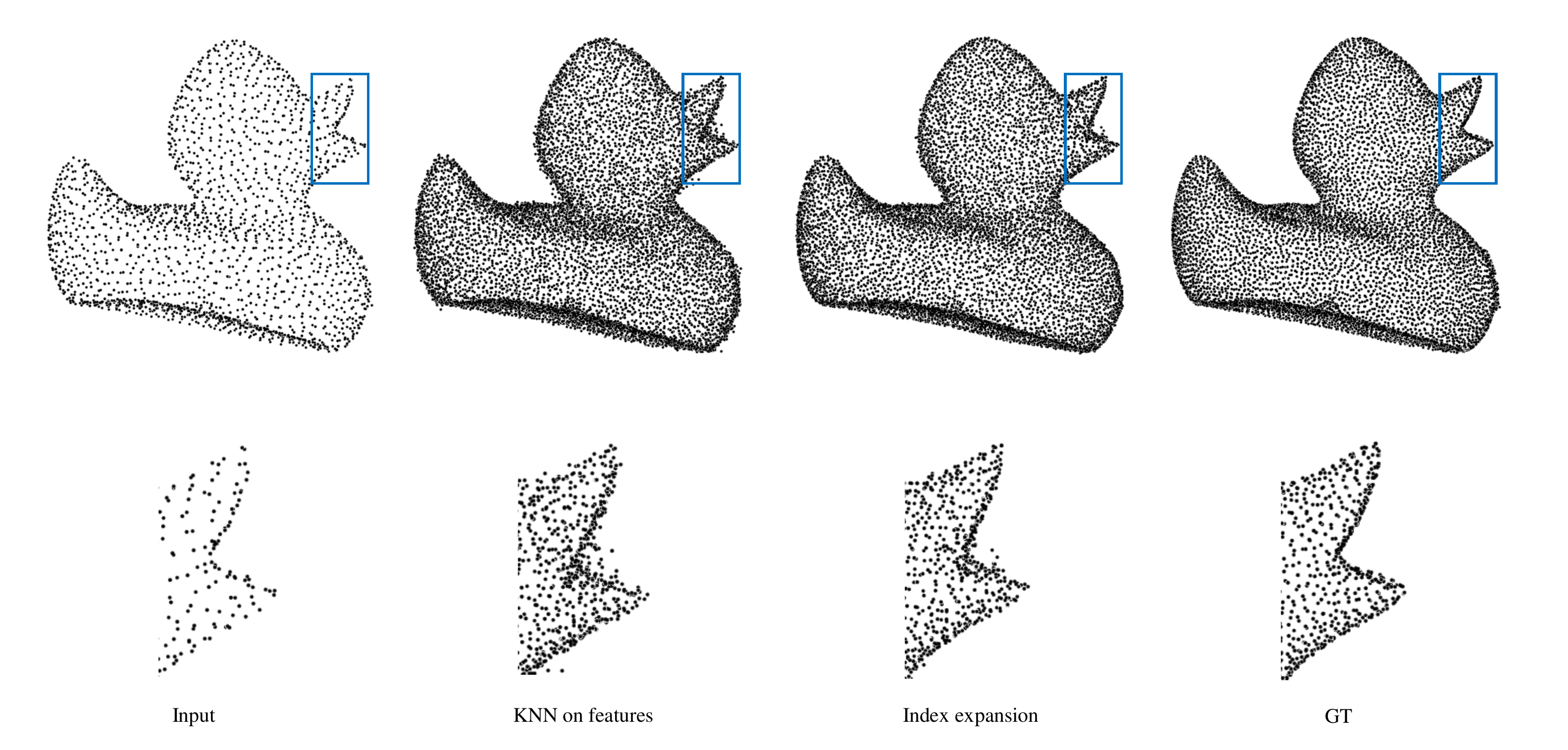}
  \caption{Visualization of different high-power index types}
  \label{fig13}
\end{figure}

We further observe the visualization results of the two types of index generation methods, as shown in the figure ~\ref{fig13}. We choose to use  more advanced PU-GCN model. The test object is a 3D duck model and the enlarged space is the corner of the mouth. This area is prone to outliers. It can be seen that using K-nearest neighbors in the feature space will generate aggregated outliers, but the outlier distance is not large. We believe that the feature k-nearest neighbors makes the generated points more distant. The mutual distance between them is closer, but it cannot be guaranteed to be close to the sparse point cloud input. Therefore, a part of the clustered outliers will be generated in the blank area for KNN on features methods. In contrast, using the point cloud K-nearest neighbor expansion method makes the generated features interact with the original point cloud position information each time the feature is upsampled, leading to fewer outliers.

Then we conduct ablation experiments for the high-power feature regression method. The point cloud upsampling model generally uses the shared parameter MLP operation at the end of model to convert the $C$-dimensional point features to 3-dimensional coordinate values. This part of the operation does not involve feature interaction between points. We propose to draw on the idea of upsampling module from image super-resolution task, performing local feature fusion on high-power point clouds. The baseline experiment is Direct regression without EdgeConv. Then we perform EdgeConv after and before coordinate regression. The quantitative results is shown in table \ref{table4}. As we can be seen from the table, in most cases, the performance of EdgeConv before point regression is optimal. Compared with direct regression, both EdgeConv is connected before and after regression will get performance improvement, especially on the more advanced PU-GCN model. Direct regression is equivalent to only using progressive upsampling, and there is no local feature fusion operation on high-power features. For the upsampling module of image super-resolution, the convolution operation is generally performed to convert channels from $C$ to 3. Similarly, the local fusion of $C$ dimension features should be performed first and then converted to three-dimensional coordinates for point cloud upsampling. Results from the talbe confirms this analogy result. After obtaining the upsampling feature of $rN\times C$, it is beneficial to generate a more uniform point cloud if the local feature fusion is performed before coordinate regression.

\begin{table}
  \centering
  \caption{Ablation study on coordinate regression type}
  \begin{tabular}{ccccc}
    \toprule
      Model&Coordinate regression type&CD\ ($10^{-3}$)&HD\ ($10^{-3}$)&P2F\ ($10^{-3}$)\\
    \midrule
    \multirow{3}{*}{PU-Net}&Direct regression&1.047&\textbf{14.311}&4.509 \\
    &EdgeConv after regression&1.013&14.884&4.686\\
    &EdgeConv before regression&\textbf{0.938}&14.512&\textbf{3.831}\\ \hline
    \multirow{3}{*}{MPU}&Direct regression&0.974&13.364&3.807 \\
    &EdgeConv after regression&0.717&11.201&\textbf{2.531}\\
    &EdgeConv before regression&\textbf{0.603}&\textbf{9.102}&2.620\\ \hline
    \multirow{3}{*}{PU-GCN}&Direct regression&0.707&10.423&2.881 \\
    &EdgeConv after regression&0.642&9.187&\textbf{2.165}\\
    &EdgeConv before regression&\textbf{0.597}&\textbf{8.138}&2.295\\
    \bottomrule
  \end{tabular}
  \label{table4}
\end{table}

\begin{figure}
  \centering
  \includegraphics[width=0.8\linewidth]{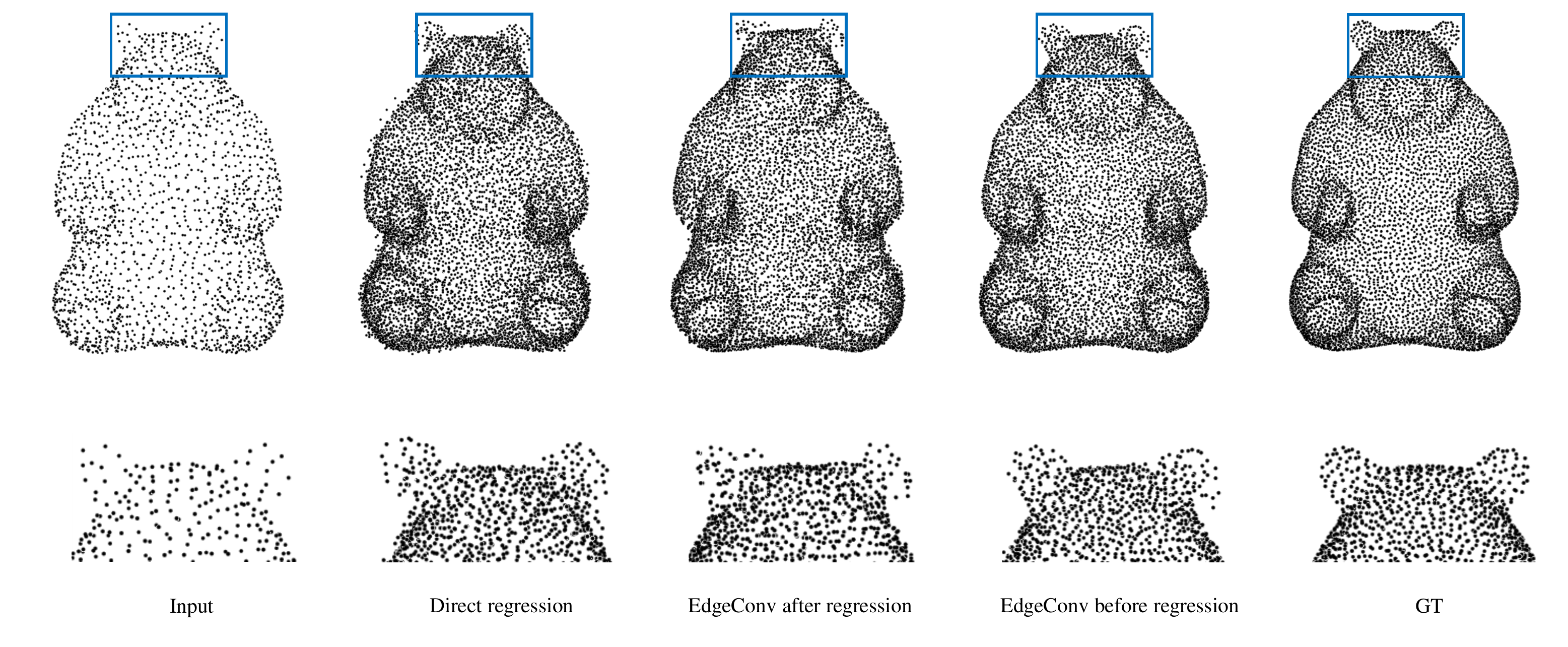}
  \caption{Visualization of different coordinate regression types}
  \label{fig14}
\end{figure}

We also presents the visualization results of this ablation experiment, as shown in the figure ~\ref{fig14}. We select MPU as base model. The test object is a 3D bear model and the enlarged space is the area of both ears. As we can be seen from the figure, the contour of the bear ears generated by performing EdgeConv before regression is closer to GT, and the overall distribution of the generated point cloud is also more uniform, which confirms the analysis above.

\section{Conclusions}
\label{sec:conclusion}
In this paper, we fairly compare the performance of various feature expansion units proposed by previous state-of-the-art point cloud upsampling models and point out that most methods fail to acquire information from neighbor point features during feature expansion process. And we propose a novel feature expansion unit named ProEdgeShuffle, which progressively fuses local feature while expanding features. Extensive experiments show that our proposed method can achieve considerable improvement over previous feature expansion units.

\bibliography{paper_data}
\end{document}